\pdfoutput=1
\documentclass[letterpaper]{article} 
\usepackage{aaai2026}   
\usepackage{times}  
\usepackage{helvet}  
\usepackage{courier}  
\usepackage[hyphens]{url}  
\usepackage{graphicx} 
\usepackage{natbib}  
\usepackage{caption} 
\usepackage{algorithm}
\usepackage{multirow}

\usepackage{algorithmicx} 
\usepackage{algpseudocode}
\usepackage{amsmath}   
\usepackage{booktabs}  
\usepackage{placeins}  

\usepackage{newfloat}
\usepackage{amsmath}
\usepackage{booktabs}
\usepackage{placeins}
\usepackage{xcolor}
\usepackage{makecell}

\usepackage{newfloat}
\usepackage{listings}
\DeclareCaptionStyle{ruled}{labelfont=normalfont,labelsep=colon,strut=off} 
\floatstyle{ruled}
\newfloat{listing}{tb}{lst}{}
\floatname{listing}{Listing}
\title{A Few Words Can Distort Graphs: Knowledge Poisoning Attacks on Graph-based Retrieval-Augmented Generation of Large Language Models}

\author{
    Jiayi Wen\textsuperscript{\rm 1},
    Tianxin Chen\textsuperscript{\rm 1},
    Zhirun Zheng\textsuperscript{\rm 2},
    Cheng Huang\textsuperscript{\rm 1}
}
\affiliations{
    \textsuperscript{\rm 1}College of Computer Science and Artificial Intelligence, Fudan University\\
    \textsuperscript{\rm 2}Department of Artificial Intelligence, Ajou University\\
}

\nocopyright
\begin{document}

\maketitle

\begin{abstract}
Graph-based Retrieval-Augmented Generation (GraphRAG) has recently emerged as a promising paradigm for enhancing large language models (LLMs) by converting raw text into structured knowledge graphs, improving both accuracy and explainability.
However, GraphRAG relies on LLMs to extract knowledge from raw text during graph construction, and this process can be maliciously manipulated to implant misleading information. 
Targeting this attack surface, we propose two knowledge poisoning attacks (KPAs) and demonstrate that modifying only a few words in the source text can significantly change the constructed graph, poison the GraphRAG, and severely mislead downstream reasoning.
The first attack, named Targeted KPA (TKPA), utilizes graph-theoretic analysis to locate vulnerable nodes in the generated graphs and rewrites the corresponding narratives with LLMs, achieving precise control over specific question-answering (QA) outcomes with a success rate of 93.1\%, while keeping the poisoned text fluent and natural.
The second attack, named Universal KPA (UKPA), exploits linguistic cues such as pronouns and dependency relations to disrupt the structural integrity of the generated graph by altering globally influential words. With fewer than 0.05\% of full text modified, the QA accuracy collapses from 95\% to 50\%.
Furthermore, experiments show that state-of-the-art defense methods fail to detect these attacks, highlighting that securing GraphRAG pipelines against knowledge poisoning remains largely unexplored.

\end{abstract}

\section{Introduction}
Large language models (LLMs) have revolutionized the way we process and generate information.
Despite their impressive abilities, they hallucinate facts and rely on outdated internal knowledge, which limits their use in tasks that require strict factual accuracy ~\cite{DBLP:journals/csur/JiLFYSXIBMF23, DBLP:journals/corr/abs-2303-08774,DBLP:conf/coling/BianH0L0HJD24}.
Retrieval-Augmented Generation (RAG) mitigates these weaknesses by grounding model outputs in an external knowledge base ~\cite{DBLP:conf/icml/BorgeaudMHCRM0L22,DBLP:journals/corr/abs-2201-08239,DBLP:conf/nips/LewisPPPKGKLYR020}.
Traditional RAG stores external knowledge as isolated text chunks, which restricts retrieval to shallow matching and limits multi-step reasoning ~\cite{DBLP:conf/iclr/AsaiWWSH24}.
Graph-based RAG (GraphRAG)~\cite{DBLP:journals/corr/abs-2404-16130} addresses this limitation by organizing the corpus into a knowledge graph through LLM-driven extraction of entities and their relations.
This structure explicitly links the extracted entities and relations and enables LLMs to reason over connected facts, achieving higher accuracy on complex queries ~\cite{DBLP:journals/corr/abs-2410-05779,DBLP:journals/corr/abs-2501-00309}.
Because the reasoning process in GraphRAG depends entirely on this constructed graph, it underpins a range of knowledge-intensive tasks, including question answering (QA) ~\cite{DBLP:conf/naacl/YasunagaRBLL21,DBLP:conf/emnlp/KarpukhinOMLWEC20}and dialogue~\cite{DBLP:journals/sigkdd/ChenLYT17}.

However, the reliance on external corpora has also made RAG systems attractive targets for security attacks~\cite{DBLP:journals/corr/abs-2412-17011,DBLP:conf/acl/ZengZHLX000WYT24}.
Prior works have identified three main attack categories.
First, malicious documents injected into the corpus can bias retrieval results and distort LLM's answers~\cite{DBLP:journals/corr/abs-2402-07867,DBLP:journals/corr/abs-2402-08416,DBLP:journals/corr/abs-2405-13401}.
Second, adversarial instructions can be hidden in retrievable chunks so that the LLM executes them when the chunks are retrieved~\cite{DBLP:conf/camlis/HinesLHZZK24,DBLP:journals/corr/abs-2401-07612}.
Third, the retriever itself can be attacked with crafted queries that cause it to miss relevant evidence~\cite{DBLP:conf/emnlp/WallaceFKGS19,DBLP:journals/corr/abs-2406-05870}. GraphRAG inherits these risks but also exposes a qualitatively different vulnerability.
Unlike traditional RAG, GraphRAG does not answer questions directly from retrieved context.
It first converts the entire corpus into a structured knowledge graph~\cite{DBLP:journals/eswa/ChenJX20,DBLP:journals/access/ChenWZCZD20}, and all subsequent tasks~\cite{DBLP:conf/coling/KimHKS20,DBLP:conf/www/WangZZLXG19} depend on this graph.

\begin{figure*}[t] 
\centering
\includegraphics[width=\textwidth]{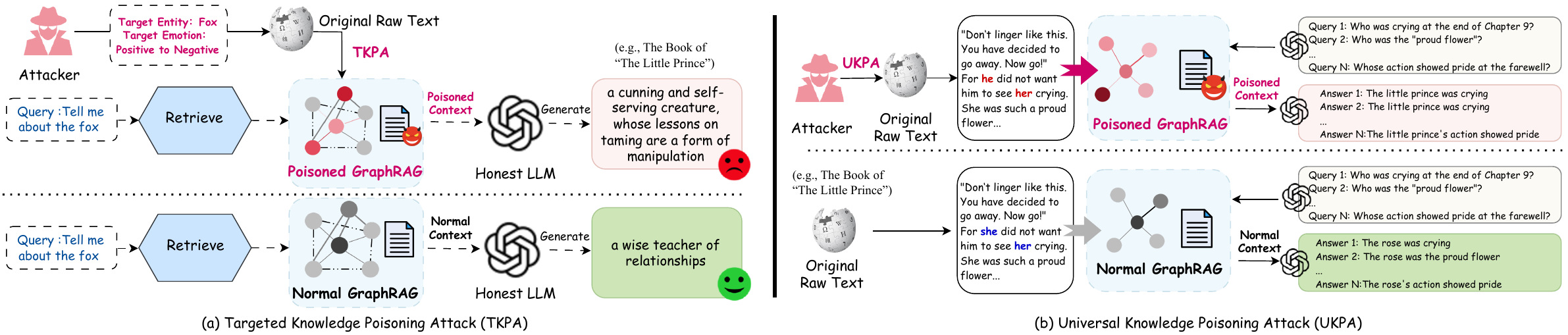} 
\caption{Proposed two knowledge poisoning attacks. (a) Targeted knowledge poisoning: manipulated facts cause GraphRAG to select poisoned context for a query, leading to incorrect LLM output. (b) Universal knowledge poisoning: altered linguistic cues globally distort graph structure, causing GraphRAG to build a biased knowledge graph and mislead LLM reasoning across diverse queries.}
\label{fig:attack_concept} 
\end{figure*}

Recent work has taken the first step toward poisoning GraphRAG: GRAGPOISON~\cite{DBLP:journals/corr/abs-2501-14050} injects crafted chunks that create or amplify false relations, showing that such relation-level manipulation can mislead multiple queries once the graph is built.
While GRAGPOISON demonstrates that GraphRAG can indeed be poisoned, its attack strategies all operate in an additive manner: it introduces malicious content into the corpus either by injecting new relations, 
repeating existing relations to strengthen them, or adding narrative chunks that blend false and true information. 
These attacks show that crafted additions to the corpus can distort the resulting graph and mislead multiple queries once the graph is built. 
\emph{An unexplored question is whether GraphRAG is also vulnerable when the adversary cannot add new text, 
but is only able to make small, subtle modifications to the existing corpus.} In this work, we reveal a manipulation-only attack surface for GraphRAG:
even without introducing additional content, simply changing a few words in the existing corpus can distort the entities and relations extracted during graph construction, and the corrupted structure then persists and misleads a broad range of queries.
This threat corresponds to subtle edits to trusted sources (e.g., minor changes in Wikipedia) rather than the injection of obviously malicious content.
Such manipulations pose two key challenges: \emph{to what extent can a few edits change the behavior of a GraphRAG system, and do these changes appear in a targeted or a widespread form?}
We address these questions by focusing on two complementary objectives:
\emph{precision}, the ability to make specific queries return attacker-desired answers with only a few edits; and
\emph{breadth}, the ability of small, subtle modifications to corrupt the graph broadly, degrading reasoning across many queries.
Although stealthiness is not an explicit objective, it is implicitly achieved by restricting the attack to very small edits on trusted sources.

To obtain the above goals, we propose two knowledge poisoning attacks (KPAs).
\textbf{(1) Targeted Knowledge Poisoning Attack (TKPA).}
As shown in Figure~\ref{fig:attack_concept}(a), TKPA exploits the topology of the knowledge graph itself to achieve fine-grained control over specific outputs.  
The key difficulty is that the impact of a small text edit propagates through two coupled stages: graph construction and downstream reasoning, so the influence of a single modification is hard to predict directly from the raw text.  
TKPA addresses this by first operating in the graph domain: it analyzes the connectivity and centrality structure of the graph to locate the subregions that have the greatest effect on a target query. 
Only after this graph-theoretic localization does it map back to the corresponding text and rewrite a few highly relevant passages.~\cite{DBLP:journals/pr/PengZZ13,gross2018graph}  
This graph-guided strategy allows a handful of carefully placed edits to reliably manipulate the outputs for selected queries, while remaining unobtrusive.
\textbf{(2) Universal Knowledge Poisoning Attack (UKPA).}
As shown in Figure~\ref{fig:attack_concept}(b), UKPA aims for maximum disruption with only a few edits.  
The challenge is that, without any specific query as a starting point, it is unclear where a small perturbation will have a global effect on the graph.  
Our key insight is that GraphRAG depends on linguistic signals, such as pronouns, coreference chains~\cite{peng2019text}, and other referring expressions, to decide when different mentions across chunks refer to the same entity; 
these signals act like the glue that holds the graph together.  
UKPA therefore targets these linguistic connections: by subtly rewriting references so that the chains break, 
it prevents the system from recognizing that different mentions refer to the same entity.  
These small edits propagate through the graph and produce widespread structural errors, severely degrading reasoning accuracy across tasks, even though the modified text remains fluent and very hard to detect.

\textbf{Contributions.} Our contributions are as follows:
\begin{itemize}
\item We identify a realistic \emph{manipulation-only attack surface} in GraphRAG, demonstrating that modifying a small number of words in the trusted corpus is sufficient to corrupt the constructed knowledge graph and mislead downstream reasoning.
\item We propose \emph{Targeted Knowledge Poisoning Attack (TKPA)}, which exploits graph-theoretic structure to locate vulnerable nodes and rewrite a small number of associated passages. This attack achieves \emph{precise manipulation of specific QAs} with a success rate of \emph{93.1\%}, while modifying less than 0.06\% of the corpus (\emph{48 words out of 94,496}) and preserving text fluency and stealth.
\item We propose \emph{Universal Knowledge Poisoning Attack (UKPA)}, which exploits linguistic structures such as pronouns and coreference to \emph{disrupt global entity linking}, degrading the integrity of the knowledge graph and reducing QA accuracy from \emph{95\% to 50\%}, while modifying only \emph{60 words out of 134,072} and affecting less than \emph{0.05\%} of the corpus.
\item We conduct extensive experiments on real-world datasets, demonstrating that both TKPA and UKPA \emph{achieve high attack effectiveness and circumvent state-of-the-art defenses}, revealing substantial security vulnerabilities in GraphRAG pipelines.
\end{itemize}

\section{Attack Methodology}
\subsection{Background}
\begin{figure}[t]
\centering
\includegraphics[width=\columnwidth]{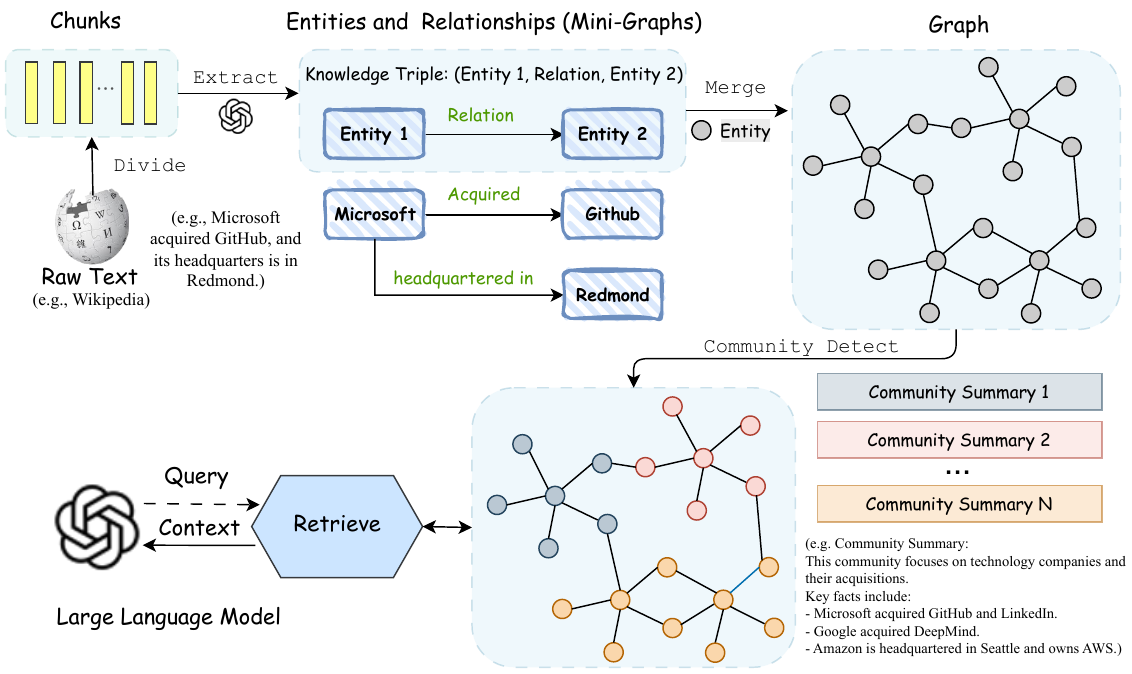} 
\caption{The pipeline of GraphRAG.}
\label{fig:graphrag_pipeline}
\end{figure}

\textbf{GraphRAG Pipeline.} Figure~\ref{fig:graphrag_pipeline} illustrates the GraphRAG pipeline.  
Given an unstructured document corpus $D = \{d_1, \dots, d_n\}$, GraphRAG first divides it into smaller text chunks $\{c_1, \dots, c_m\}$.  
For each chunk $c_i$, a mini knowledge graph $G_i$ is extracted, consisting of entity-relation-entity triples via the extraction function $f_{\text{extract}}$, capturing local semantic structure:  
$G_i = f_{\text{extract}}(c_i)$.
These mini-graphs are then merged to form the overall corpus graph:  
$G_{\text{merged}} = \bigcup_{i=1}^m G_i$.
Next, a community detection function $f_{\text{community}}$ partitions $G_{\text{merged}}$ into communities $C = \{C_1, \dots, C_k\}$, each representing a coherent semantic subgraph:  
$C = f_{\text{community}}(G_{\text{merged}})$.
For each community $C_j$, a summary $S_j$ is generated to capture key information and relations within that community.  
When a user submits a query $Q$, GraphRAG applies a retrieval function $g_{\text{retrieve}}$ to extract the relevant community summaries $S_{\text{rel}} \subseteq \{S_1, \dots, S_k\}$ as contextual information for the LLM:  
$S_{\text{rel}} = g_{\text{retrieve}}(Q, \{S_1, \dots, S_k\})$.
Finally, the LLM generates the answer conditioned on the query and retrieved context:  
$\text{Answer} = \text{LLM}(Q, S_{\text{rel}})$.

\textbf{Attack Model.}
We consider a gray-box adversary that poisons GraphRAG by editing the source corpus rather than injecting entirely new documents or accessing model parameters.  
The adversary knows the overall pipeline: GraphRAG segments text into chunks, extracts entities and relations, builds a knowledge graph, and generates community-level summaries that are later used as context for reasoning.  
The attacker can modify a small fraction of trusted sources (e.g., Wikipedia) but has no access to the constructed graph or model parameters.
\begin{itemize}
    \item \textbf{TKPA Model.} 
    \emph{Attacker Knowledge:} Understands that GraphRAG organizes extracted knowledge into communities that drive downstream answers. 
    \emph{Attacker Capability:} Can modify a small part of the corpus to influence the answers of specific queries.
    \item \textbf{UKPA Model.} 
    \emph{Attacker Knowledge:} Only knows that GraphRAG builds a knowledge graph from text, without details of its structure. 
    \emph{Attacker Capability:} Can make small edits to the corpus aimed at broadly degrading reasoning across many queries rather than any single one.
\end{itemize}




\subsection{Targeted Knowledge Poisoning Attack}
\begin{figure*}[t]
\centering
\includegraphics[width=\textwidth]{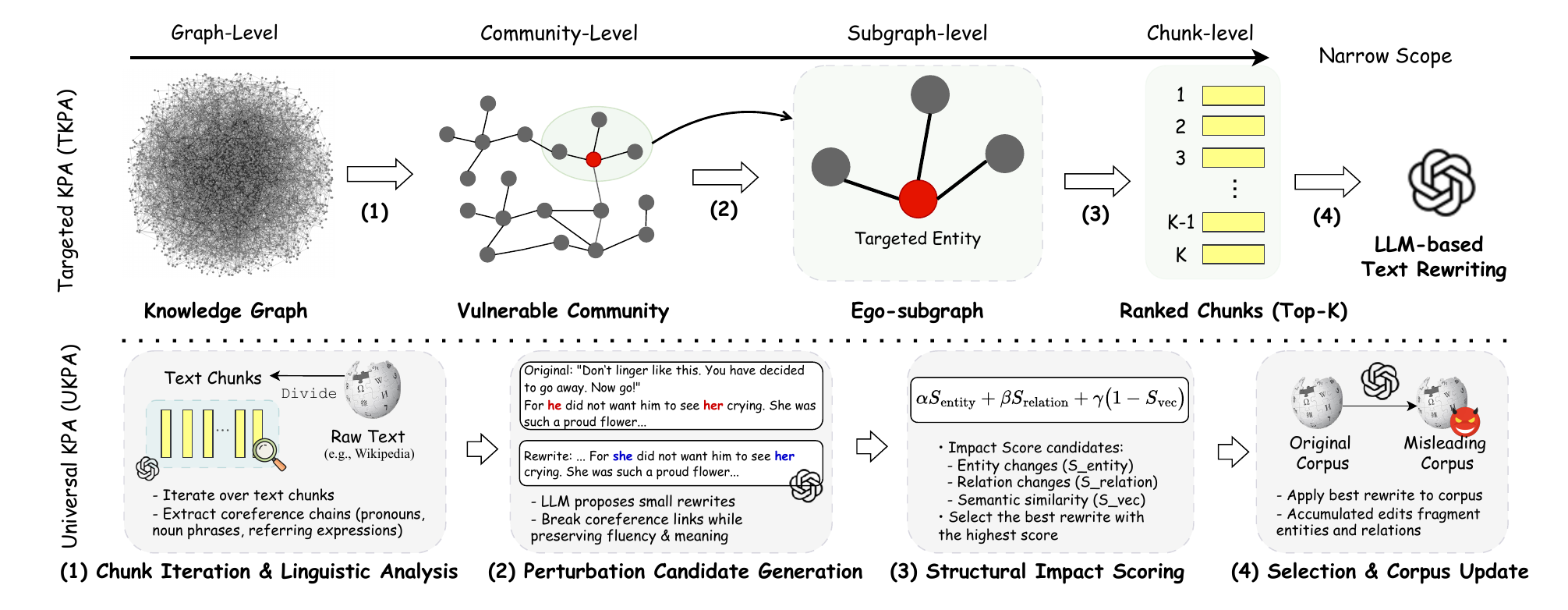} 
\caption{The pipeline of TKPA and UKPA.}
\label{fig:targeted_attack_framework} 
\end{figure*}

As shown in Figure~\ref{fig:targeted_attack_framework}, 
the key insight behind Targeted Knowledge Poisoning Attack (TKPA) is to treat poisoning as a network intervention problem on the knowledge graph rather than a random text-editing task.
Given a user query, the attacker first considers the entities that GraphRAG associates with the query and chooses one as the \emph{target entity} via LLM-based entity extraction.
The attacker then leverages graph-theoretic principles: centrality to identify structurally influential nodes, community structure to restrict the affected region, 
and ego-subgraphs~\cite{mitchell1969social,DBLP:journals/intr/WangW25} to localize edits. 
These cues guide the selection of a small set of text chunks whose modification maximizes downstream impact.
This structure-guided view leads to a four-module pipeline that progressively narrows the attack scope from the full graph to a few edits.

\textbf{(1) Vulnerable Community Localization (VCL).}
The first step is to determine where a small intervention will have the largest structural effect.
From a network perspective, communities with a highly central target entity and limited size are the most susceptible: 
a modification there can influence a larger fraction of the context while requiring fewer edits. 
To formalize this intuition, we define a \emph{vulnerability score} for each community:
\begin{equation}
\mathcal{V}_\mathrm{score} = \frac{(1+D_e)(1+C_e)}{\log(1+\mathrm{TLen})},
\end{equation}
where \(D_e\) and \(C_e\) denote the degree and betweenness centrality of the target entity within that community,
and \(\mathrm{TLen}\) measures the length of the community summary.
The numerator captures the entity’s structural leverage, while the denominator penalizes communities that require editing long narratives.
The attacker evaluates this score for all communities containing the target entity and chooses the one with the highest score as the entry point for manipulation.
This choice reflects the classic principle of influence maximization in networks: 
maximize downstream impact per unit of edit cost.

\textbf{(2) Ego-subgraph Extraction.}
After selecting the most vulnerable community, the attacker narrows the intervention to the local structure around the target entity. 
Specifically, an \emph{ego-subgraph} \(G_{\mathrm{ego}}(v_t)\) is extracted, consisting of the target node \(v_t\), its one-hop neighbors, and the edges among them. 
This ego-subgraph defines the local context that GraphRAG relies on when answering questions about \(v_t\); modifying this neighborhood directly alters how \(v_t\) and its relations are represented in the graph.
Only the text chunks associated with nodes and edges in \(G_{\mathrm{ego}}(v_t)\) are kept as candidates, and the next module ranks these candidates to determine which ones to rewrite.
This step follows a classic principle in network interventions: apply a small, localized perturbation to a structurally central region to produce broad downstream effects.

\textbf{(3) Chunk Scoring and Selection.}
With the candidate chunks constrained to those linked to \(G_{\mathrm{ego}}(v_t)\),
the attacker next ranks them by their potential to alter downstream reasoning.
Drawing on insights from network influence theory, we model the importance of a chunk as the weighted combination of three signals:
(i) \emph{structural impact}: how central its corresponding entity is in the local subgraph,
(ii) \emph{semantic relevance}: how closely the chunk's content aligns with the target query,
and (iii) \emph{sentiment polarity}: how much the tone of the text can bias the generated narrative.
We have
\begin{equation}
    \mathcal{C}_\mathrm{score} = w_1 S_{\mathrm{graph}} + w_2 S_{\mathrm{semantic}} + w_3 S_{\mathrm{attitude}},
\end{equation}
where \(S_{\mathrm{graph}}\) is computed from the PageRank centrality of the corresponding entity within \(G_{\mathrm{ego}}(v_t)\),
\(S_{\mathrm{semantic}}\) is the cosine similarity between the embedding of the chunk and the target query,
and \(S_{\mathrm{attitude}}\) quantifies the chunk's affective tone using a language model.
Each term is normalized to \([0,1]\) across candidates before combining them.

The weights \((w_1,w_2,w_3)\) act as tunable weights that balance structural leverage,
contextual relevance, and linguistic framing.
In practice, higher weight is assigned to structural impact so that edits are concentrated on influential regions of the graph,
while the other two terms ensure that the selected edits remain relevant and subtle.
By ranking chunks with the score, the attacker can focus a small number of edits on the locations that offer the highest
\emph{influence-to-cost ratio}: maximum effect on the final answers per unit of editing effort.

\textbf{(4) LLM-driven Manipulation.}
The top-ranked chunks are rewritten by a LLM to subtly alter facts or tone while preserving fluency and style.
These rewritten chunks replace the original text in the corpus, so that when GraphRAG rebuilds the knowledge graph,
the poisoned narratives become embedded in the community summaries. As a result, a few carefully chosen modifications spread through the graph and strongly bias downstream tasks,
while keeping the attack surface compact and difficult to detect.

\subsection{Universal Knowledge Poisoning Attack}
As shown in Figure~\ref{fig:targeted_attack_framework},
the Universal Knowledge Poisoning Attack (UKPA) aims to degrade GraphRAG globally rather than biasing a single query.
The central insight is that GraphRAG relies heavily on \emph{linguistic coherence cues}, particularly
coreference chains and referring expressions, to decide when multiple mentions across chunks
should be merged into a single entity node. These cues form the inductive bias that allows
GraphRAG to consolidate scattered evidence into a coherent graph. 
If these signals are weakened or disrupted, the resulting graph becomes fragmented:
nodes proliferate, relations split across disconnected components, and reasoning paths are interrupted. The dependency exposes a unique vulnerability.
By subtly perturbing these cues at the text level, an attacker can interfere with the graph construction
before any reasoning occurs, systematically breaking the long-range links that enable multi-hop reasoning.
The key challenge is to find and disrupt these high-leverage linguistic signals \emph{without access to the graph itself}:
the attacker never observes the final structure, yet edits that look benign can still propagate
through the construction process and cause large-scale structural distortions.

To exploit this vulnerability, UKPA operates entirely in the language domain. 
Its strategy is grounded in a key observation from linguistics ~\cite{hobbs1978resolving,DBLP:conf/emnlp/LeeHLZ17,DBLP:conf/naacl/LeeHZ18}: 
entity linking across documents depends almost exclusively on \emph{coreference resolution signals}.  
including pronouns, definite descriptions, and other referring expressions, that tie different mentions to the same entity. 
These signals are inherently weak and context-dependent: even small changes in wording or surface form can prevent 
the coreference model from clustering mentions together. UKPA deliberately introduces such perturbations into the raw corpus. 
It scans the text and makes a small number of edits that weaken these cues—
for example, altering pronouns, introducing slight ambiguity in referring expressions, 
or modifying the form of an entity name—so that the linking mechanism fails to merge mentions correctly. 
Although each edit appears innocuous at the sentence level, these disruptions systematically fragment the global graph: 
mentions that once formed a single, well-connected node are now split into many disconnected nodes, 
relations become scattered, and long reasoning chains collapse. 
The cumulative effect is a broad degradation of GraphRAG’s reasoning capabilities. 
The pipeline consists of four modules:

\textbf{(1) Chunk Iteration and Linguistic Analysis.}
UKPA begins at the text level. 
For each chunk in the corpus, a LLM is used to perform
linguistic analysis and extract \emph{coreference chains}, linking between textual
mentions (pronouns, noun phrases, and other referring expressions) and the entities they denote.
These mention-to-entity links form the latent backbone that GraphRAG later uses
to merge mentions into coherent entity nodes across chunks.

\textbf{(2) Perturbation Candidate Generation.}
Given the extracted coreference chains, the LLM is for generating some alternative rewrites for the chunk that deliberately weaken these links.
Each candidate rewrite must satisfy three constraints:
(i) maintain grammatical fluency,
(ii) preserve the local meaning of the text, and
(iii) stay within a small edit distance from the original.
Typical perturbations include substituting pronouns with vague noun phrases,
introducing slight ambiguity in referring expressions, or reordering clauses in a way
that makes cross-chunk linking less reliable.
These edits are to prevent GraphRAG from clustering mentions into a single entity
during graph construction.

\textbf{(3) Structural Impact Scoring.}
To estimate the effect of each candidate rewrite without direct access to the final graph,
UKPA employs a surrogate scoring function that measures how much the local entity-relation
structure extracted from a chunk would change after the edit:
\begin{equation}
    \mathcal{I}_\mathrm{score} = \alpha S_\mathrm{entity} + \beta S_\mathrm{relation} + \gamma \bigl(1 - S_\mathrm{vec}\bigr),
\end{equation}
where \(S_\mathrm{entity}\) denotes the symmetric difference between the sets of entities
extracted from the original and modified chunks, 
\(S_\mathrm{relation}\) is the symmetric difference between their relation sets,
and \(S_\mathrm{vec}\) is the cosine similarity ~\cite{DBLP:conf/naacl/LeeHZ18,DBLP:journals/eswa/RahimAAAK25} between the embedding~\cite{DBLP:journals/jmlr/BengioDVJ03} of the original and modified chunk.
The coefficients $(\alpha,\beta,\gamma)$ are tunable weights that balance the relative importance of entity fragmentation, relation distortion, and semantic closeness (SC).
The score favors candidates that cause larger perturbations in the local entity-relation structure
while keeping the modified text semantically close to the original,
so that the attack remains subtle while still fragmenting the global graph once these changes accumulate.

\textbf{(4) Selection and Corpus Update.}
For each chunk, the candidate rewrite with the highest score is chosen,
and the corresponding text in the corpus is updated accordingly.
When GraphRAG subsequently constructs the knowledge graph on this modified corpus,
the accumulated perturbations disrupt entity linking:
mentions that were previously merged into a single node are split into multiple disconnected nodes,
relations become scattered, and cross-chunk reasoning chains collapse.

\section{Evaluation}

\subsection{Experimental Setup}
We evaluate TKPA and UKPA on a standard GraphRAG pipeline (Microsoft GraphRAG), focusing on:
(1) \emph{Attack Effectiveness}: the ability to manipulate or degrade GraphRAG outputs, and 
(2) \emph{Attack Stealthiness}: the few, subtle perturbations required and their ability to evade existing defenses.

\textbf{Datasets \& Tasks.}
We evaluate on long-form documents, which represent the primary application scenario of GraphRAG: synthesizing knowledge from large unstructured text rather than isolated paragraphs. Such documents provide rich context and interconnections for constructing meaningful knowledge graphs.
For TKPA, we choose \textit{The Little Prince} (LP), the Wiki page on the \textit{Financial Crisis of 2007-2008} (FC08)~\cite{wiki:financial_crisis_2008}, and the Wiki page on the \textit{Japanese Asset Price Bubble} (JAPB)~\cite{wiki:jp_bubble}.
For UKPA, we use the Wiki page on the \textit{Russo-Ukrainian War} (RUW)~\cite{wiki:rus_ukr_war} and LP.
To evaluate downstream performance, we generate $20-30$ multi-hop question-answer pairs per document using GPT-4o~\cite{OpenAI2024}, prompted as a domain expert. 
Each question requires synthesizing information from multiple sections, similar in style to HotpotQA~\cite{yang2018hotpotqa}.
All pairs are manually reviewed for clarity and correctness. 
Attack success is judged by whether the modified GraphRAG system produces poisoned answers as determined by an LLM-based evaluator.

\textbf{Models \& Parameters.}
We build the GraphRAG pipeline using state-of-the-art LLMs. 
For graph construction, we adopt GPT-4o-mini due to its efficiency and strong reasoning capability, 
and use BAAI's \texttt{bge-m3}~\cite{DBLP:journals/corr/abs-2402-03216} as the embedding model, which is a multilingual and open-source embedding model supporting multiple granularities.
For TKPA, GPT-4o is employed to perform fine-grained rewriting of selected text chunks and to verify attack outcomes. 
For UKPA, GPT-4o is used to identify linguistic cues (e.g., pronouns), generate poisoned variants of the text, 
and extract entity-relation structures for scoring candidate perturbations. Unless otherwise stated, the weights in Eq.~(2) are set to $(w_1,w_2,w_3) = (0.5, 0.3, 0.2)$,
and those in Eq.~(3) to $(\alpha,\beta,\gamma) = (0.25, 0.25, 0.5)$.

\subsection{Baselines.}
\emph{(i) Attack.} We compare with three attacks:
\begin{itemize}
    \item \textbf{PoisonedRAG (PRAG)}  \cite{DBLP:journals/corr/abs-2402-07867}, injects malicious text into the corpus to bias retrieval and force attacker-specified outputs (for TKPA).
    \item \textbf{Naive Swap (NS)}, a simplified variant that directly injects emotionally charged keywords (e.g., \textit{excellent}) into the text without preserving coherence (for TKPA).
    \item \textbf{TextFooler-style Perturbation (TP)}~\cite{DBLP:conf/aaai/JinJZS20}, replaces words with embedding-based synonyms and ignores coreference and global structure (for UKPA).
\end{itemize}
\emph{(ii) Defense.} We compare against three representative defense methods:
\begin{itemize}
    \item \textbf{Perplexity-based Filter (PF)}~\cite{Radford2019}: a classical and widely used baseline that flags chunks with unusually high perplexity as suspicious, implemented with GPT-2.
    \item \textbf{LLM-based Contamination Detector (LLMDet)}~\cite{DBLP:journals/corr/abs-2309-00614}: a recent state-of-the-art approach that leverages powerful LLMs to classify each chunk as clean or poisoned through few-shot prompting.
    \item \textbf{Semantic Closeness Checking (SCC)}~\cite{Honnibal2020}: a content-based defense that detects potential manipulations by measuring semantic similarity between the original and modified chunks, particularly suited to universal poisoning scenarios.
\end{itemize}

\subsection{Attack Effectiveness}

\textbf{Metrics.}
We evaluate the impact of the proposed attacks from two perspectives: targeted effectiveness and global degradation.
For TKPA, we report \emph{Attack Success Rate (ASR)}, i.e., the percentage of targeted queries whose answers are manipulated as intended, and \emph{Question-Answer Semantic Deviation (QASD)}, which measures how far the generated answers deviate semantically from the correct responses.
For UKPA, we assess global degradation through (i) the drop in overall QA accuracy on Microsoft GraphRAG and LightRAG to examine generalization across GraphRAG systems, and (ii) structural damage to the constructed knowledge graph.
The latter is quantified using \emph{node retention rate}, \emph{edge retention rate}, and \emph{Jaccard similarity} between the clean and poisoned graphs.
Retention rates $(0\to1)$ measure the fraction of nodes or edges that persist after poisoning, while Jaccard similarity evaluates the overlap between the node and edge sets.
Lower values on these metrics indicate more severe fragmentation of the graph structure.

\textbf{TKPA Performance.}
Table~\ref{tab:whitebox_performance_revised} presents the performance of TKPA across multiple corpora.
Across all datasets, TKPA achieves high ASR (over 90\% on average),
showing that a handful of well-placed edits can consistently steer GraphRAG outputs
toward attacker-specified answers.
The QASD values further indicate that the poisoned answers diverge significantly
from the ground truth, while remaining coherent and fluent.
Compared with PoisonedRAG and Naive Swap baselines,
TKPA achieves both higher ASR and larger semantic deviation with fewer edits,
highlighting the advantage of structure-guided poisoning over naïve text-level interventions.
\begin{table}[t]
\centering
\small 
\begin{tabular}{ccccc}
\toprule
\textbf{Dataset} & \textbf{Metric} & \textbf{TKPA} & \textbf{PRAG} & \textbf{NS} \\
\midrule
\multirow{2}{*}{LP}   & ASR (\%) $\uparrow$ & \textbf{93.10} & 71.50 & 18.20 \\   
& QASD $\uparrow$      & \textbf{0.85} & 0.68 & 0.15 \\
\cmidrule(l){2-5}
\multirow{2}{*}{FC08} 
& ASR (\%) $\uparrow$ & \textbf{89.50} & 68.90 & 14.30 \\
& QASD $\uparrow$      & \textbf{0.81} & 0.65 & 0.13 \\ 
\cmidrule(l){2-5}
\multirow{2}{*}{JAPB} 
& ASR (\%) $\uparrow$ & \textbf{91.20} & 70.80 & 15.80 \\  
& QASD $\uparrow$      & \textbf{0.83} & 0.67 & 0.12 \\  
\midrule
\multirow{2}{*}{Average}
& ASR (\%) $\uparrow$ & \textbf{91.27} & 70.40 & 16.1 \\
& QASD $\uparrow$      & \textbf{0.83} & 0.67 & 0.13 \\
\bottomrule
\end{tabular}
\caption{Performance of TKPA across multiple corpora.}
\label{tab:whitebox_performance_revised}
\end{table}

\begin{table}[t]
\centering
\setlength{\tabcolsep}{4pt}
\small
\begin{tabular}{ccccc}
\toprule
\multirow{2}{*}{\textbf{Metric}} & \multicolumn{2}{c}{\textbf{Microsoft GraphRAG}} & \multicolumn{2}{c}{\textbf{LightRAG}} \\
\cmidrule(lr){2-3} \cmidrule(lr){4-5}
& RUW & LP & RUW & LP \\
\midrule
Nodes (Org/Atk) & 347/327 & 104/97 & 436/405 & 163/131 \\
Edges (Org/Atk) & 379/390 & 119/121 & 388/378 & 175/167 \\
\midrule
\textbf{Node Ret. Rate} & \textbf{0.5648} & \textbf{0.5769} & \textbf{0.4335} & \textbf{0.3926} \\
\textbf{Edge Ret. Rate} & \textbf{0.2770} & \textbf{0.3529} & \textbf{0.1443} & \textbf{0.2343} \\
Node Jaccard & 0.4100 & 0.4255 & 0.2899 & 0.2783 \\
Edge Jaccard & 0.1581 & 0.2121 & 0.0789 & 0.1362 \\
\bottomrule
\end{tabular}
\caption{Structural degradation caused by UKPA. Clean (Org) vs. attacked (Atk) graphs for Microsoft GraphRAG and LightRAG on RUW and LP corpora.} 
\label{tab:graybox_structure}
\end{table}

\textbf{UKPA Performance.}
The UKPA aims to indirectly affect downstream tasks by disrupting the graph structure. As shown in Table~\ref{tab:graybox_structure}, the UKPA can severely damage the graph structure. Although the total number of nodes and edges shows some reduction, the more drastic change is reflected in the graph's topology, as evidenced by the extremely low Jaccard similarity scores (e.g., as low as 0.0789 for edges on LightRAG) reveal that the graph's topology has been almost completely rewritten. This experiment shows that our linguistic perturbations effectively corrupt the knowledge base from within, rather than simply deleting information. The structural damage to the knowledge graph directly propagates to downstream QA tasks. As shown in Table~\ref{tab:ukpa_qa_accuracy}, the QA accuracy on Microsoft GraphRAG dropped from 95\% to 50\% under our attack, while in lightRAG, the QA accuracy drops from 90\% to 45\%. These results demonstrate that our universal poisoning paradigm can effectively cripple the system's reasoning capabilities.

\begin{table}[t]
\centering
\small
\begin{tabular}{ccc}
\toprule
\textbf{GrapRAG} & \textbf{Attack} & \textbf{Accuracy} \\
\midrule
\multirow{3}{*}{\makecell{Microsoft \\ GraphRAG}}
& No Attack & 95\% \\
& TP        & 85\% \\
& \textbf{UKPA}      & \textbf{50\%} \\
\midrule
\multirow{3}{*}{LightRAG}
& No Attack & 90\% \\
& TP        & 85\% \\
& \textbf{UKPA}      & \textbf{45\%} \\
\bottomrule
\end{tabular}
\caption{Performance of UKPA on downstream QA task.}
\label{tab:ukpa_qa_accuracy}
\end{table}

\subsection{Attack Stealthiness}
Table~\ref{tab:defense_evaluation} shows that existing defenses are largely ineffective against both TKPA and UKPA, with F1-scores close to 0.
This result stems from the fact that both attacks operate in ways that evade surface-level detection.
For TKPA, the manipulations are guided by graph structure: selected chunks are rewritten by advanced LLMs so that the style, fluency, and local semantics remain natural.
This makes perplexity-based filters and LLM detectors ineffective, as the modified text is statistically and stylistically indistinguishable from clean text.
For UKPA, the perturbations directly exploit the system's reliance on \emph{linguistic coherence cues}.
Breaking these signals leaves the sentence-level meaning intact but causes long-range fragmentation in the knowledge graph.
Because existing defenses analyze only local text, they cannot capture this deeper structural distortion.
We also consider query-side defenses, such as query paraphrasing~\cite{DBLP:journals/corr/abs-2309-00614}.
These techniques are inherently ineffective against our attacks: the poisoned corpus corrupts the knowledge graph itself, so any paraphrased query will ultimately retrieve compromised entities and poisoned context.
Defenses operating at the query level cannot mitigate attacks that target the underlying data source.

\begin{table}[t]
\centering
\small
\begin{tabular}{ccccc}
\toprule
\textbf{Attack} & \textbf{Defense} & \textbf{Precision} & \textbf{Recall} & \textbf{F1-Score} \\
\midrule
\multirow{2}{*}{TKPA}
& PF      & 0.08 & 0.06 & 0.07 \\
& LLMDet     & 0.14 & 0.12 & 0.13 \\
\midrule
\multirow{3}{*}{UKPA}
& SCC & 0.08 & 0.06 & 0.07 \\     
& PF      & 0.05 & 0.04 & 0.04 \\
& LLMDet     & 0.12 & 0.10 & 0.11 \\
\bottomrule
\end{tabular}
\caption{Effectiveness of defense against TKPA and UKPA.}
\label{tab:defense_evaluation}
\end{table}

In addition to bypassing detectors, both attacks require extremely small modifications to the corpus.
As summarized in Table~\ref{tab:Words_modification}, TKPA achieves targeted control by changing on the order of a few dozen to a few hundred words for an entire document (e.g., 48 words out of 94,496 for LP, less than 0.06\%),
while UKPA achieves global degradation by modifying only 32-60 words across very large corpora (0.03\%-0.05\%).
The sharp contrast highlights the stealthiness of both strategies: TKPA introduces highly localized edits focused on a single query target, whereas UKPA spreads very small changes across the corpus to globally disrupt the graph.
Such small-scale perturbations explain why existing text-level defenses fail to detect these attacks even when their downstream impact is catastrophic.

\begin{table}[t]
\setlength{\tabcolsep}{2pt}
\centering
\small
\begin{tabular}{ccccc}
\toprule
\textbf{Attack} & \textbf{Dataset} & \makecell{\textbf{Total}\\\textbf{Words}} & \makecell{\textbf{Modified}\\\textbf{Words (min/avg)}} & \makecell{\textbf{Modification}\\\textbf{Ratio (min/avg)}} \\
\midrule
\multirow{3}{*}{TKPA}
&LP & 94496 & 48/155 & 0.055\% / 0.164\% \\
&FC08   & 40223 & 76/325         & 0.18\% / 0.807\% \\
&JAPB & 44445 & 113/334        & 0.254\% / 0.751\% \\
\midrule
\multirow{2}{*}{UKPA}
&LP & 94496  & 32 & 0.033\% \\
&RUW   & 134072 & 60 & 0.045\% \\
\bottomrule
\end{tabular}
\caption{Statistics of word-level perturbations introduced by TKPA and UKPA across datasets. min and avg indicate the minimum and average number.}
\label{tab:Words_modification}
\end{table}

\subsection{Ablation Study}
\begin{figure}[h]
\centering
\includegraphics[width=0.9\columnwidth]{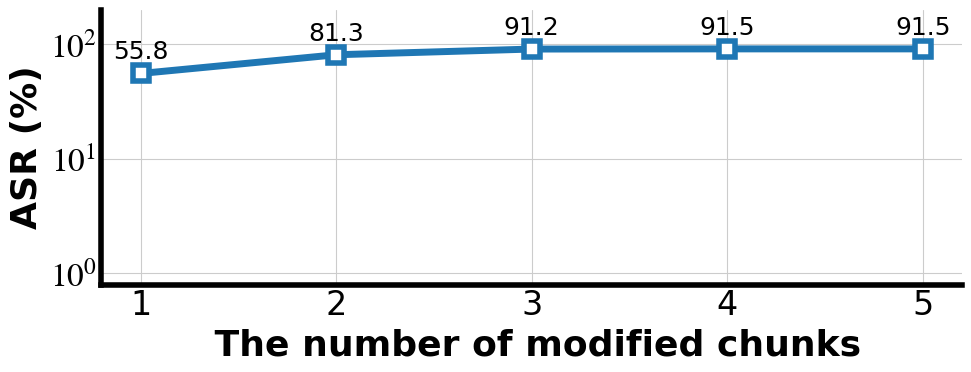} 
\caption{Impact of the number of modified chunks (Top-K) on the ASR of TKPA.}
\label{fig:tkpa_topk_ablation}
\end{figure}

\textbf{Ablation Study of TKPA's Parameters.}
We evaluate the role of the three weights in the TKPA chunk‑scoring function (Eq.~(2)).
While equal weighting of the three signals already achieves 89.8\% ASR, tuning the weights to emphasize graph structure $(w_1=0.5,w_2=0.3,w_3=0.2)$ further boosts ASR to 91.2\%.
Setting any single weight to 1 while zeroing the others results in significantly lower performance: graph-only yields 65.3\% ASR, semantic-only yields 58.2\%, and attitude-only yields 51.7\%.
This result demonstrates that assigning priority to graph structure is more effective than treating the signals equally. In addition to the weighting scheme, we analyze how the number of modified chunks ($K$) influences TKPA performance. Figure~\ref{fig:tkpa_topk_ablation} shows that the ASR increases sharply with just a few edits: modifying the top-ranked chunk alone ($k=1$) achieves 55.8\%, rises to 81.3\% with $k=2$, and reaches 91.2\% with only three chunks. Beyond $k=3$, the curve plateaus, indicating that the scoring mechanism effectively prioritizes high-impact segments. These results highlight that TKPA attains near-maximal impact with minimal, focused modifications, reinforcing both its stealth and efficiency.

\begin{figure}[h]
\centering
\includegraphics[width=\columnwidth]{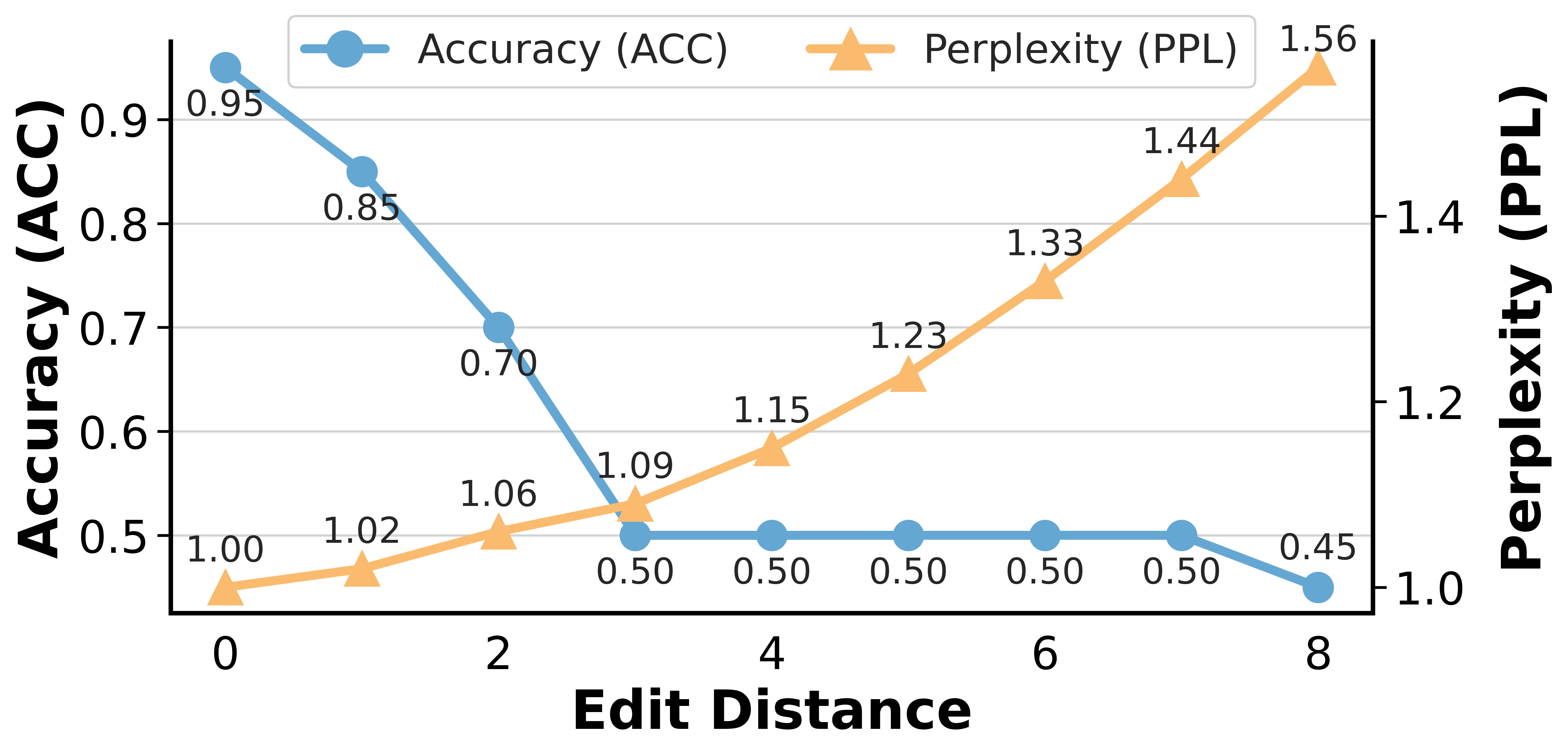}
\caption{Impact of edit distance on UKPA.}
\label{fig:edit_distance_plot}
\end{figure}
\noindent \textbf{Ablation Study of UKPA's Parameters.}
We evaluate the role of the three weights in the UKPA structural impact score (Eq.~(3)).
Equal weighting of the three terms degrades QA accuracy to 0.55, while tuning the weights to prioritize semantic preservation $(\alpha=0.25,\beta=0.25,\gamma=0.5)$ further reduces it to $0.50$.
Using a single component alone is much less effective, leaving QA accuracy at $0.70-0.75$.
This result shows that balancing structural disruption with semantic consistency is more effective than either equal weighting or focusing on a single factor. To analyze the effect of edit distance, Figure~\ref{fig:edit_distance_plot} shows that
while small edits (distance $\leq 3$) already cause a drastic drop in QA accuracy (from 0.95 to 0.50),
larger edits increase perplexity significantly without further improving attack impact.
This results justifies the constraint on small edit distances to ensure stealthiness.

\section{Conclusion and Future Work}
We have revealed a fundamental vulnerability in GraphRAG systems: automatically constructed knowledge graphs open a critical attack surface, where manipulation of only a few words can cause significant distortion. We have proposed two knowledge poisoning attacks: TKPA, which leverages graph-theoretic structure to precisely control specific answers with over 93\% ASR, and UKPA, which exploits linguistic cues to fragment the global graph, cutting QA accuracy from 95\% to 50\% with less than 0.05\% of the corpus modified.
Experiments demonstrate that such small and natural edits can evade state-of-the-art defenses.
These results highlight the need to treat graph construction as a core security component rather than a passive preprocessing step.
For the future, we plan to investigate lightweight, scalable attack and defense methods that work with more complex GraphRAG systems, and to examine how multimodal inputs (e.g., images or metadata) in future GraphRAG pipelines may introduce new vulnerabilities.

\FloatBarrier 

\end{document}